\def\paperID{8033}
\def\confName{ICCV}
\def\confYear{2025}

\def\authorBlock{
    Xiang Fang\textsuperscript{1}\thanks{Equal contribution} \qquad
    Shihua Zhang\textsuperscript{1}\footnotemark[1]  \qquad
	Hao Zhang\textsuperscript{1} \qquad
	Tao Lu\textsuperscript{2} \qquad
	Huabing Zhou\textsuperscript{2} \qquad
     Jiayi Ma\textsuperscript{1}\thanks{Corresponding author. E-mail address: jyma2010@gmail.com}\footnotemark[2]\\
     \textsuperscript{1} Wuhan University \quad \textsuperscript{2} Wuhan Institute of Technology\\
}

\newif\ifreview 
\newif\ifarxiv \newcommand{\arxiv}{\arxivtrue}
\newif\ifcamera 
\newif\ifrebuttal 

\arxiv 

\pdfoutput=1
\documentclass[10pt,twocolumn,letterpaper]{article}
\ifreview \usepackage[review]{cvpr} \fi
\ifarxiv \usepackage[pagenumbers]{cvpr} \fi
\ifrebuttal \usepackage[rebuttal]{cvpr} \fi
\ifcamera \usepackage{cvpr} \fi

\usepackage{graphicx}	
\usepackage{amsmath}	
\usepackage{amssymb}	
\usepackage{booktabs}
\usepackage{times}
\usepackage{microtype}
\usepackage{epsfig}
\usepackage{caption}
\usepackage{float}
\usepackage{placeins}
\usepackage{color, colortbl}
\usepackage{stfloats}
\usepackage{enumitem}
\usepackage{tabularx}
\usepackage{xstring}
\usepackage{multirow}
\usepackage{xspace}
\usepackage{url}
\usepackage{subcaption}
\usepackage{xcolor}
\usepackage[hang,flushmargin]{footmisc}
\usepackage{amsmath}
\usepackage{multirow}
\usepackage{graphicx}
\usepackage{subcaption}
\usepackage{mathrsfs}
\usepackage{color}
\usepackage{amssymb}
\usepackage{pifont}
\usepackage{svg}
\usepackage{enumerate}
\usepackage{colortbl}
\usepackage{listings}
\usepackage{mathtools}

\ifcamera \usepackage[accsupp]{axessibility} \fi

% ...

\ifarxiv  \fi

\newcommand{\R}[1]{{%
    \textbf{%
        \ifstrequal{#1}{1}{\textcolor{red}{R#1}}{%
        \ifstrequal{#1}{2}{\textcolor{blue}{R#1}}{%
        \ifstrequal{#1}{3}{\textcolor{magenta}{R#1}}{%
        \ifstrequal{#1}{4}{\textcolor{teal}{R#1}}{%
                           \textcolor{cyan}{R#1}%
        }}}}%
    }%
}}

\usepackage{xr-hyper}

\makeatletter
\newcommand*{\addFileDependency}[1]{
  \typeout{(#1)}
  \@addtofilelist{#1}
  \IfFileExists{#1}{}{\typeout{No file #1.}}
}

\makeatother

\definecolor{cvprblue}{rgb}{0.21,0.49,0.74}
\usepackage[pagebackref,breaklinks,colorlinks,allcolors=cvprblue]{hyperref}
\usepackage[capitalize]{cleveref}
\crefname{section}{Sec.}{Secs.}
\crefname{table}{Table}{Tables}
\crefname{figure}{Fig.}{Figs.}

\ifarxiv \crefname{appendix}{App.}{Apps.}
\else \crefname{appendix}{Suppl.}{Suppls.} \fi

\begin{document}
\title{Selecting and Pruning: A Differentiable Causal Sequentialized State-Space Model\\ for Two-View Correspondence Learning}
\author{\authorBlock}
\maketitle

\begin{abstract}
Two-view correspondence learning aims to discern true and false correspondences between image pairs by recognizing their underlying different information. 
Previous methods either treat the information equally or require the explicit storage of the entire context, tending to be laborious in real-world scenarios. 
Inspired by Mamba's inherent selectivity, we propose \textbf{CorrMamba}, a \textbf{Corr}espondence filter leveraging \textbf{Mamba}'s ability to selectively mine information from true correspondences while mitigating interference from false ones, thus achieving adaptive focus at a lower cost.
To prevent Mamba from being potentially impacted by unordered keypoints that obscured its ability to mine spatial information, we customize a causal sequential learning approach based on the Gumbel-Softmax technique to establish causal dependencies between features in a fully autonomous and differentiable manner. 
Additionally, a local-context enhancement module is designed to capture critical contextual cues essential for correspondence pruning, complementing the core framework. 
Extensive experiments on relative pose estimation, visual localization, and analysis demonstrate that CorrMamba achieves state-of-the-art performance. Notably, in outdoor relative pose estimation, our method surpasses the previous SOTA by $2.58$ absolute percentage points in AUC@20\textdegree, highlighting its practical superiority. Our code will be publicly available.
\end{abstract}

\section{Introduction}
\label{sec:intro}

\begin{figure}[t]
  \centering
  \includegraphics[width=0.99\linewidth]{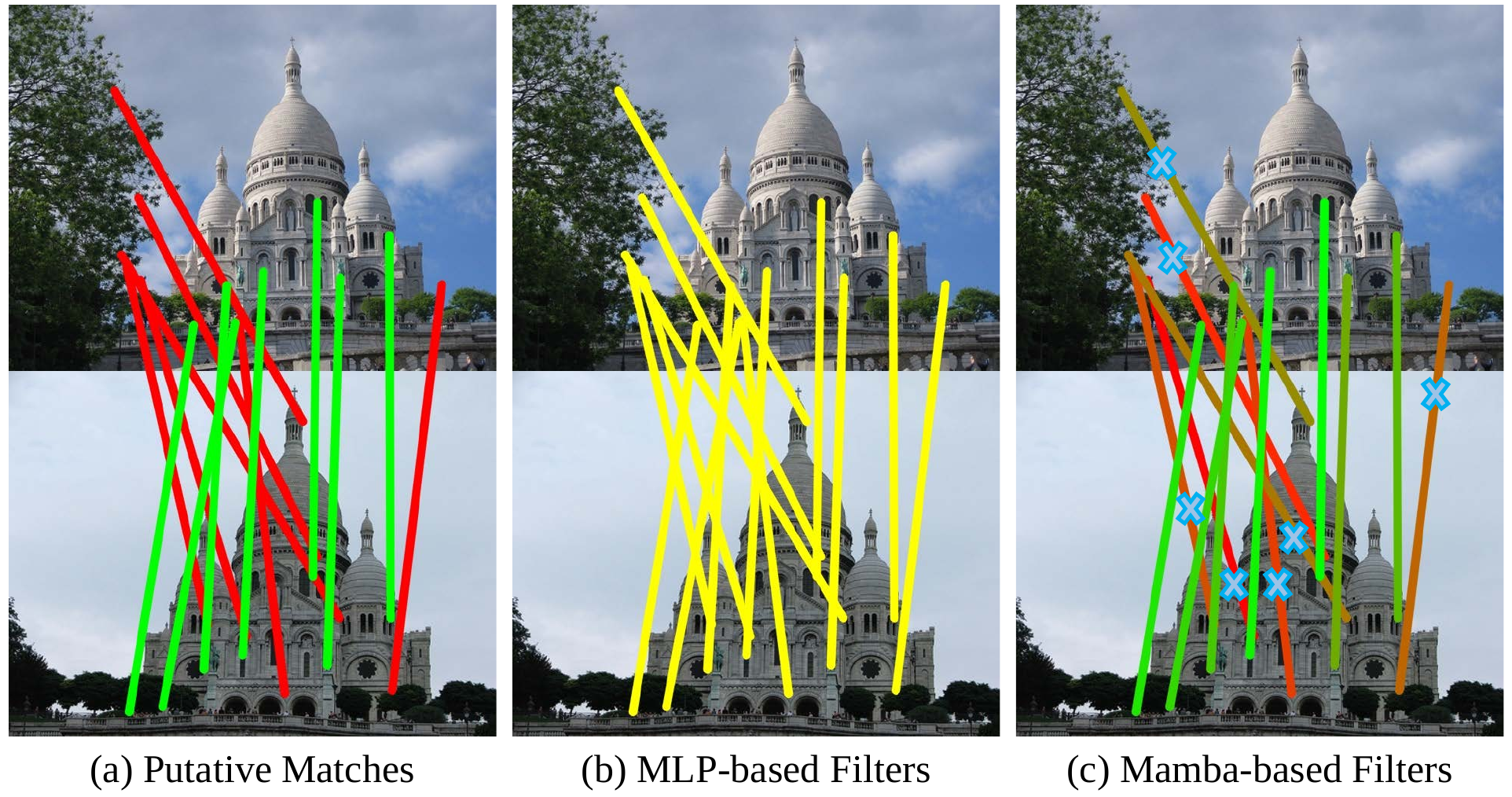}
  \vspace{-0.08in}
  \caption{Illustration of different filters of linear complexity for correspondence learning. False matches are shown in red (\textcolor[rgb]{1,0,0}{—}), and correct ones in green (\textcolor[rgb]{0,1,0}{—}). The transition of the yellow hue from closer to red to closer to green in (c) signifies a progression from lower to higher weights, meaning the model is more likely to consider the match as an inlier and ignore outliers at the same time.}
  \label{illustration}
\end{figure}


Two-view correspondence learning that finds sparse matches and estimates geometric relationships for image pairs is a fundamental task and perennial theme in computer vision. It is of paramount importance for many downstream tasks such as image retrieval~\cite{tolias2016image} and Structure from Motion~\cite{saputra2018visual}. The typical pipeline is divided into two stages, namely generating a putative correspondence set and removing false matches (\emph{i.e.} outliers)~\cite{ma2021image}. However, constrained by the limited discriminative ability of descriptors, the sparse and irregular matches in the putative set usually contain a large number of outliers, and removing them is imperative for image matching. Therefore, in this paper, we focus on outlier rejection to maintain true correspondences (\emph{i.e.} inliers), which facilitates accurately estimating the true model.

In accordance with previous studies, nascent outlier rejection methods~\cite{fischler1981random, ma2014robust} 
have demonstrated good performance in situations with a modicum of outliers. Woefully, they often fail in real-world scenarios with numerous outliers. Among the burgeoning learning-based approaches, PointCN~\cite{yi2018learning} and some subsequent works~\cite{zhang2019learning, zhao2021progressive} begin to regard the correspondence pruning as a binary classification problem. Yet, as shown in \cref{illustration}(b), they employ multi-layer perceptrons (MLPs) that treat each input equally, which fails to adequately discern the differences in information resulting from inliers and outliers. PointACNe~\cite{sun2020acne} customizes to give different attention to diverse inputs through explicit weights, but this simplistic approach is insufficient for handling complex scenarios. Additionally, some vanilla attention-based methods~\cite{liu2023progressive, li2024mc, fang2024acmatch} modulate the importance of information by constructing feature maps for the inputs and assigning different weights. 
Albeit with commendable performance, 
they are not particularly efficient as they require explicitly storing the entire context, directly leading to significant computational resource consumption during training and inference~\cite{gu2023mamba}.
Therefore, we consider the following question:  \emph{How to selectively treat inliers and outliers, and fully capture the information while compressing the context into a more compact representation as much as possible?}

Of late, a novel architecture termed Selective State Space Model, \emph{i.e.} Mamba~\cite{gu2023mamba}, has gained widespread adoption in visual tasks~\cite{liu2024vmamba, zhu2024vision}. One of its key advantages is \emph{selectivity}, enabling models to focus on or disregard specific inputs. 
Furthermore, it achieves optimization of complexity through compressing context into a compact representation and employing hardware scanning methods. 
Inspired by this, we seek to introduce Mamba to tackle the problem of two-view correspondence learning. However, most existing Mamba-based visual models are tailored for tasks involving regular images as input~\cite{guo2024mambair,zhu2024vision}, and their common structures are ill-suited for two-view correspondence learning with sparse points. Although some recent endeavors have explored using Mamba for point cloud tasks~\cite{liu2024point, zhang2024point}, they predominantly focus on manually designing causal sequences and lack guidance on how to adapt to our 2D data, which are usually unordered and contain a high proportion of outliers, while inliers being subject to geometric consistency constraints. 

For the aforesaid purpose, we propose a Mamba-based correspondence learning method termed CorrMamba, which is superior in terms of effective data selection. 
Specifically, we first meticulously design a differentiable causal sequence learning block (CSLB) inspired by the Gumbel-Softmax technique~\cite{jang2017categorical}. This enables the data fed into Mamba to maintain certain dependencies, ensuring that the geometric estimation remains unaffected by input permutations, \emph{i.e.} permutation invariance~\cite{zhang2019learning}. 
Additionally, it's known that a critical prerequisite in correspondence learning is the ability to capture local context~\cite{yi2018learning}. Therefore, to increase the capability to obtain context information, we design a Local Graph Pattern Learning (LGPL) module.
The Channel-Aware Mamba filter (CAMF) that follows is the core part, which can adaptively adjust the parameters according to the input.
This allows the network to distinguish inliers and outliers in the high-dimensional latent space and give different focus, which is beneficial to correspondence pruning.
 As shown in \cref{illustration}(c), our Mamba-based approach can differentiate between true and false correspondences tellingly, and ignore false ones at the same time. This kind of selective focus does not require our model to explicitly store the entire context, as is common with attention-based methods, thus achieving context compression.

In a nutshell, the contributions of this study mainly include the following three folds: 
{\bf (i)} To our best knowledge, this represents the inaugural exploration into leveraging Mamba for sparse feature matching. Additionally, we pioneer the application of Mamba's selection traits to tackle scenarios characterized by a significant presence of outliers adeptly. This design could potentially offer valuable insights for a variety of other applications;
{\bf (ii)} We innovatively customize a differentiable sequence construction strategy, which transforms the unordered features into a causal sequence in a fully autonomous and differentiable manner, facilitating Mamba's learning of geometric relationships. 
{\bf (iii)} We design a suite of experiments to substantiate the rationality of our methodological design and further demonstrate the efficacy and robust generalization capabilities of our approach through various real-world datasets.

\section{Related Work}
\label{Related Work}

\subsection{Correspondence Learning}
A renowned paradigm involves first obtaining an initial set of correspondences, followed by the application of outlier rejection to refine and achieve a more accurate correspondence set.
The best-known handcrafted methods such as RANSAC~\cite{fischler1981random} and its variants~\cite{torr2000mlesac, chum2005matching} employ a hypothesis-verification strategy to find a maximal consistent subset that fits a particular geometric model. This type of method is highly dependent on the sampled subsets, leading to their failure in high outlier situations. 
Then, with the development of deep learning, MLPs are first used as a solution for correspondence pruning. PointCN~\cite{yi2018learning} utilizes them to extract high-dimensional features of each putative correspondence individually and introduces context normalization to capture global contextual information. OANet~\cite{zhang2019learning} introduces a differentiable layer that captures local information by softly assigning nodes to a set of clusters. LMCNet~\cite{liu2021learnable} combines global and local coherence to robustly detect true correspondences. ConvMatch~\cite{zhang2023convmatch} uses a CNN as the backbone and avoids the design of additional context normalization modules of MLP-based approaches. However, these methods treat each input equally and lack the ability to mine the different information between inliers and outliers.
Some subsequent approaches~\cite{liu2023progressive, li2024mc, zhang2024dematch} attempt to address the divergence of information between outliers and inliers by learning an attention map that softly assigns varying degrees of focus to different regions. However, they suffer from explicitly storing the entire context, directly leading to significant computational resource consumption.

Another popular pipeline involves directly obtaining an accurate set of correspondences in one go. Representative work such as SuperGlue~\cite{sarlin2020superglue} has made significant strides by utilizing graph neural networks in conjunction with attention mechanisms~\cite{vaswani2017attention}. However, these methods experience a quadratic decrease in efficiency as the number of keypoints increases, leading to difficulties in practical application. Detector-free dense matching methods~\cite{edstedt2023dkm,edstedt2024roma} and semi-dense approaches~\cite{sun2021loftr} can achieve accurate matches even in extreme scenarios like textureless regions, but the increased computational cost and memory usage due to richer matches remain unresolved. Therefore, our paper focuses on the first paradigm, while the second pipeline is discussed appropriately in the analysis (see \cref{matcher}).

\subsection{State Space Models}
The state space model (SSM) is initially used to describe dynamic systems. 
The problem of early SSMs is similar to RNN models that easily forget global contextual information and suffer from gradient vanishing as the sequence length grows~\cite{gu2021combining}. Recent work Mamba~\cite{gu2023mamba} based on SSM successfully overcomes these shortcomings. 
With the subquadratic complexity and selective scanning mechanism, Mamba has the potential to be a prospective backbone. 
 While Vision Mamba~\cite{zhu2024vision} and VMamba~\cite{liu2024vmamba} make innovative applications of Mamba architectures to raw visual inputs through their token-wise processing paradigm, their inherent design for sequential data structures inherently conflicts with the spatially structured nature of our coordinate-based point-wise tasks. Furthermore, Point Mamba~\cite{liu2024point} and PCM~\cite{zhang2024point} pioneer applying Mamba to 3D sequences, which exhibit closer structural affinity with our data configurations. However, these approaches inadequately leverage Mamba's inherent selective processing capabilities and rely on manually engineered data serialization schemes that may not generalize across different data types. To this end, we conduct in-depth investigations into developing a fully automated strategy for learning sequential patterns in input data. Moreover, we propose an enhanced Mamba Filter that leverages its inherent selective characteristics to both extract and utilize inliers' information while simultaneously rejecting outliers.

\section{Preliminaries}
\label{Preliminaries}
\subsection{Revisiting Mamba}
Mamba~\cite{gu2023mamba} originates from the SSMs, which are initially employed for linear time-invariant systems to map the input $x(t)\in \mathbb{R}$ to the output $y(t)\in \mathbb{R}$~\cite{kalman1960new}:
\begin{equation}
    \dot{h}(t) =Ah(t)+Bx(t),  \quad  y(t) =Ch(t),
    \label{shi1}
\end{equation}
where $h(t)\in \mathbb{R}^N$ is the hidden state, $\dot{h}(t)\in \mathbb{R}^N$ is the derivative of the hidden state. $A\in \mathbb{R}^{N\times N}$, $B\in \mathbb{R}^{N\times 1}$, and $C\in \mathbb{R}^{1\times N}$ are the parameters of the model. The continuous-time SSM can be discretized to a discrete-time SSM~\cite{gu2023mamba} by zero-order hold (ZOH) discretization as:
\begin{equation}
    h_{k}=\bar{A}h_{k-1}+\bar{B}x_{k},\quad y_{k}=\bar{C}h_{k},
    \label{eq:shi2}
\end{equation}
where $\bar{A}$, $\bar{B}$ and $\bar{C}$ are discrete forms of $A$, $B$, and $C$. 
Building on this foundation, Mamba can incorporate a selection mechanism that allows the parameters to be changed from fixed to a function of the inputs while changing the tensor shape. We use linear layers denoted as:
\begin{equation}
    s_B(x),s_C(x), s_{\Delta}(x)\!=\!\operatorname{Linear}_B(x),\operatorname{Linear}_C(x),\operatorname{Linear}_{\Delta}(x),
    \label{5}
\end{equation}
where $s_B(x)$, $s_C(x)$, and $s_{\Delta}(x)$ are the parameters of $B$, $C$, and the step size $\Delta$ with regard to the input $x$, respectively.

\subsection{Problem Formulation}
\label{Problem}
Given an image pair $(\mathbf{I},\mathbf{I'})$, we employ off-the-shelf feature detectors to extract keypoints. Subsequently, employing the nearest neighbor (NN) method, we establish initial putative matches. Denoted as $\mathbf{C} = [\mathbf{c}_1;\mathbf{c}_2;\ldots]$, where $\mathbf{c}_i=\{(x_i, y_i, x_i',y_i')|i = 1, \ldots, N\}$, $(x_i,y_i)$ and $(x_i',y_i')$ represent the coordinates of the $i$-th keypoints in respective images. 
In analogy to \cite{yi2018learning}, we approach the two-view correspondence learning problem as an inlier/outlier classification and an essential matrix regression. For the putative correspondences $\mathbf{C}\in\mathbb{R}^{N\times 4}$, to extract the deep information, they are usually upscaled to get $\mathbf{F}=\{\mathbf{f}_1,\mathbf{f}_2,\ldots,\mathbf{f}_N\}\in \mathbb{R}^{N\times d}$ 
using  MLPs. Moreover, we establish an inlier predictor at each layer of the network for simultaneous training, with only the last predictor yielding the probability value $\mathbf{P} = [p_1, p_2, \ldots, p_N ]^T \in \mathbb{R}^{N\times1}$, where $p_i\in[0,1)$ signifies the probability that the corresponding $\mathbf{c}_i$ is an inlier. Like other learning-based methods~\cite{liu2021learnable,li2024mc}, we employ a weighted eight-point method based on $\mathbf{P}$ to directly estimate the essential matrix. The entire process is encapsulated as:
\begin{equation}
\begin{aligned}
\widehat{\mathbf{P}}&=f_{\phi}({\mathbf{C}}), ~\
\widehat{\mathbf{E}}&=g(\widehat{\mathbf{P}},{\mathbf{C}}),
\end{aligned}
\end{equation}
where $f_{\phi}(\cdot)$ undertakes inlier prediction, and $g(\cdot)$ signifies the estimation of the parametric model.

\section{Methodology}
\label{Methodology}
\begin{figure*}[t]
\centering
	\includegraphics[width=\linewidth]{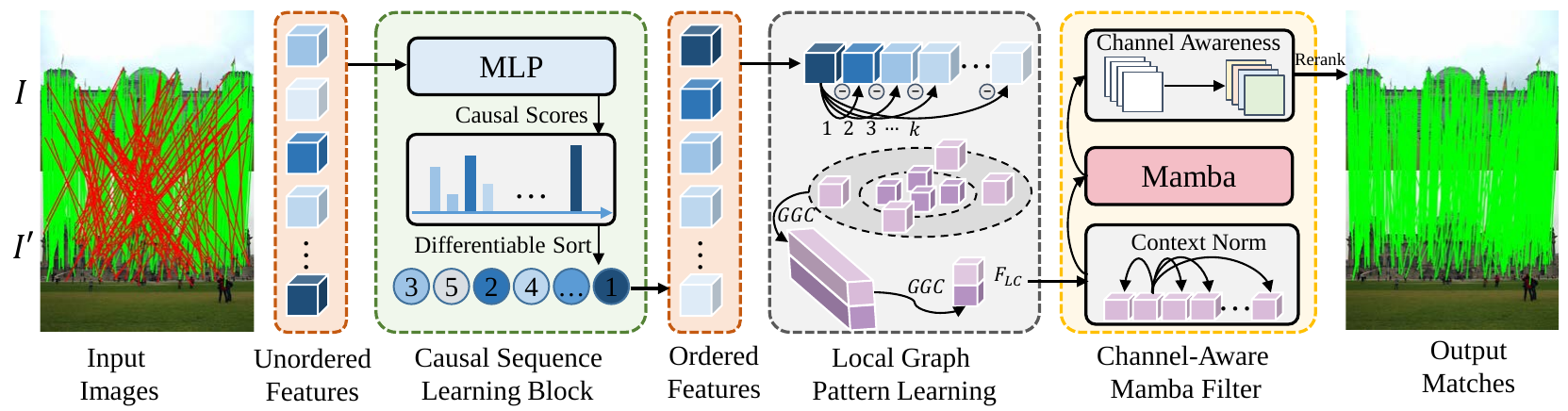}
    \vspace{-0.1in}
	\caption{Framework diagram of CorrMamba. We use the putative set of correspondences obtained by off-the-shelf detectors as input and finally obtain the inlier probability of each correspondence through the network.}
	\label{framework}
\end{figure*}

In this chapter, we propose the first mamba-inspired feature matching method named CorrMamba. As illustrated in \cref{framework}, our approach processes unordered features through the Causal Sequence Learning Block (CSLB) in \cref{sec:cslb} to autonomously and differentiably learn causal sequences. Subsequently, the ordered features undergo local context enhancement via the Local Graph Pattern Learning (LGPL) module in \cref{sec_LCEM} and selective information mining through the Channel-Aware Mamba Filter (CAMF) in \cref{sec_mamba}. The output features are positional reranked and passed through an Inlier Predictor to obtain the final matches. 

Our network incorporates fundamental components including model estimation, Order-Aware Blocks, and inlier predictors similar to 
NCMNet~\cite{liu2023progressive} and BCLNet~\cite{miao2024bclnet}, which are not fully drawn for visual clarity.
Subsequent sections will systematically dissect the main parts.

\subsection{Causal Sequence Learning}
\label{sec:cslb}
The intrinsic complexity of keypoints’ spatial arrangements poses challenges for direct application of Mamba~\cite{gu2023mamba}. Specifically, naively inputting raw feature representations compromises the model’s capacity to extract geometric correspondences, as Mamba’s architecture is inherently optimized for sequential data with distinctive causal dependencies (e.g., text or audio). To address this challenge, we propose a Causal Sequential Learning Block (CSLB) that explicitly models the causal dependencies among spatially distributed visual elements. 
While existing sequence construction methods~\cite{zhu2024vision,zhang2024point} rely on \emph{handcrafted} scanning mechanisms, which inherently conflict with the spatial-geometric coherence of unordered feature points through artificial ordering constraints, our CSLB achieves dependency learning in a data-driven manner. 
Specifically, we propose a learnable Scorer that formulates geometrically meaningful relationships between features $\mathbf{f}_i$:
\begin{equation}
    s_i = \operatorname{Scorer}(\mathbf{f}_i),  
    \label{scorer}
\end{equation}
which consists of convolutions, normalizations, and activation functions, as depicted in the \emph{supplementary material}. 

Given that our scoring system $S=\{s_1, s_2, \ldots, s_{N}\}$ is entirely learned autonomously without explicit human interventions, we establish a feature sorting mechanism based on the learned scores to achieve optimal alignment with Mamba's preference. 
However, sorting features based on the scores is not directly amenable to optimization via gradient descent methods. To this end, we incorporate the Gumbel-Softmax technique~\cite{jang2017categorical}, which formulates a differentiable approximation by reparameterization. For each $s_i$, we introduce Gumbel noise, which injects randomness into the sampling process: 
\begin{equation}
    y_i \!=\! \operatorname{softmax}[(g_i+\operatorname{log} s_i)\!/\! \tau]
        \!=\! \frac{\operatorname{exp}((g_i+\operatorname{log} s_i)/\tau)}{\sum_1^{N}\! \operatorname{exp}((g_i+\operatorname{log} s_i)\!/\!\tau)},
\end{equation}
where $g_i=-\operatorname{log}(-\operatorname{log}(u_i))$ denotes Gumbel noise term associated with $s_i$, $u_i$ follows a standard uniform distribution, and $\tau\in\mathbb{R}^+$ serves as a temperature parameter modulating the trade-off between distribution sharpness and output smoothness. 

The obtained $\mathcal{Y} = \{y_1, y_2, \ldots, y_{N}\}$ can be regarded as a set of soft values, which are amenable to gradient propagation. Departing from conventional Gumbel-Softmax, we employ Top-N instead of Top-K, ensuring that we rank all the features:
\begin{equation}
    \mathcal{I} = \operatorname{Top-K}(S,k=N),\quad Y = \operatorname{sort}(\mathcal{Y},\mathcal{I}).
\end{equation}
Note that $\operatorname{sort}(\cdot, \cdot)$ accepts the argument $\mathcal{I}$ to performs index-based sorting of $\mathcal{Y}$. This means instead of just sorting in ascending or descending order, $\mathcal{Y}$ is indexed according to $\mathcal{I}$.

We proceed by generating a sequence of hard values $\hat{Y} = \{\hat{y}_1, \hat{y}_2, \ldots, \hat{y}_{N}\}=\{1,1,\ldots, 1\}$ with $\mathcal{I}$ (the value 1 is set according to the index present in $\mathcal{I}$), which can aid in the sorting of features: 
\begin{equation}
    R = \hat{Y}-\operatorname{detach}(Y)+Y,
\end{equation}
where the detach operation is employed to prevent gradients from propagating in specific directions. Consequently, the resulting $R$ has the same numerical values as $\hat{Y}$ and the same gradients as $Y$. This means that $R$ equals 1 when backpropagated and we simply use it for gradient backpropagation. 
We utilize $R$ and $\mathcal{I}$ to sort the features:
\begin{equation}
    \mathbf{F}'=\operatorname{sort}(\mathbf{F},\mathcal{I})\cdot R. 
    \label{F_11}
\end{equation}

Drawing inspiration from the differentiable Top-K selection approach of Gumbel-Softmax, we implement a fully autonomous and differentiable method to gain the ordered feature $\mathbf{F}'=\{\mathbf{f}'_1, \mathbf{f}'_2, \ldots, \mathbf{f}'_{N}\}$.

 \subsection{Local Graph Pattern Learning}
 \label{sec_LCEM}
Having established the causal sequence requirement for Mamba's inputs, we now turn our attention to its second fundamental limitation: restricted local context modeling. While the Mamba architecture~\cite{gu2023mamba} has global information aggregation capabilities, existing approaches~\cite{hatamizadeh2024mambavision,wang2024mamba} typically require auxiliary designs to address this critical limitation in local feature extraction -- a crucial capability for feature matching tasks~\cite{zhang2019learning}. To address this gap, we propose a Local Graph Pattern Learning (LGPL) module to enhance local context modeling capabilities.  
 To effectively leverage local geometric relationships, we formulate a dynamic neighborhood graph $\mathcal{G}_i=\{\mathcal{V}_i,\mathcal{E}_i\}$ in the feature space for each correspondence $\mathbf{f}'_i$, where vertex set $\mathcal{V}_i=\{\mathbf{f}'_{i1},\cdots,\mathbf{f}'_{ik}\}$ comprises the $k$-nearest neighbors based on spatial adjacency, and edge set $\mathcal{E}_i=\{e_{i1},\cdots,e_{ik}\}$ represents directed feature-space connections. Adapting the established edge construction strategy similar to~\cite{liu2023progressive}, we define edges as:
 \begin{equation}
     e_{ij}=[\mathbf{f}'_i || \mathbf{f}'_i-\mathbf{f}'_{ij}],j=1,2,\cdots,k,
 \end{equation}
 where $[\cdot || \cdot]$ denotes channel-wise concatenation operator.

To enable structured learning of local graph patterns, we develop a group-wise graph convolution (GGC) mechanism with hierarchical normalization. The proposed method operates through three key phases. Firstly, nodes in $\mathcal{V}_i$ are partitioned into $g$ distinct groups based on their feature similarity to the anchor $\mathbf{f}'_i$, with each group containing $\frac{k}{g}$ topologically adjacent nodes. This grouping establishes $\mathcal{G}_i^1,\cdots,\mathcal{G}_i^g$ subgraphs that preserve local geometric structures. Secondly, We design two convolutions with kernel $1\times \frac{k}{g}$ and $1\times g$ for $\mathcal{E}_i$ to obtain $F_{GGC}$. In addition, to incorporate more context information, we continue to increase the context-aware capability of graph patterns:  
 \begin{equation}
     \mathbf{F}_{LC}=\delta(\operatorname{BN}(\operatorname{IN}(\mathbf{F}_{GGC}))),
     \label{14}
 \end{equation}
 where $\operatorname{IN}{(\cdot)}$ and $\operatorname{BN}{(\cdot)}$ denotes Instance Normalization and Batch Normalization.

\subsection{Channel-Aware Mamba Filter}
\label{sec_mamba}
Building upon the enhanced local features $\mathbf{F}_{LC}$, we develop a selective correspondence filtering mechanism based on Mamba~\cite{gu2023mamba}. Our key insight stems from the observation that inliers and outliers exhibit fundamentally different information characteristics. We need to strategically mine and effectively utilize the information from these diverse points to better prune correspondences. To this end, leveraging Mamba's inherent selectivity, we design a Mamba-based filter, which could focus on the inputs from inliers while disregarding outlier information and compressing the deep features into a more compact representation. 

\begin{figure}[t]
  \centering
  \includegraphics[width=0.7\linewidth]{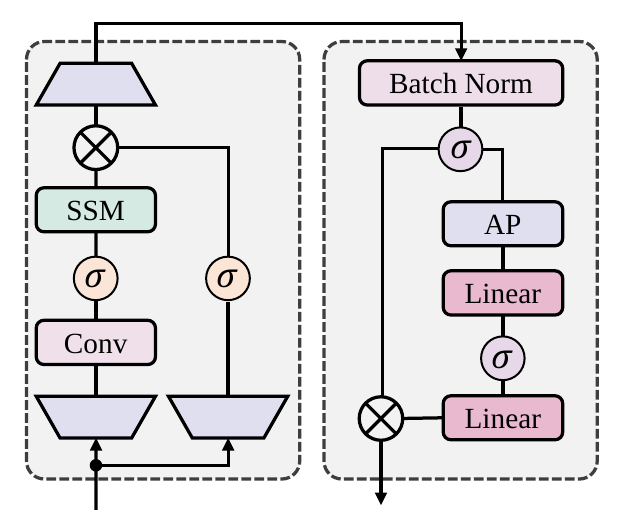}
  \vspace{-0.1in}
  \caption{Architecture of Channel-Aware Mamba Filter. Note that residual connections have been omitted for visual clarity.}
  \label{mamba}
\end{figure}

First of all, as shown in the left half of \cref{mamba}, we input $\mathbf{F}_{LC}$ into Mamba to obtain global information, which can be expressed as: 
\begin{equation}
    {\mathbf{F}_M=\mathbf{F}_{LC}+\mathcal{M}_1(\operatorname{SSM}(\operatorname{CNN}(\mathcal{M}_2(\mathbf{F}_{LC})))\odot \delta(\mathcal{M}_3(\mathbf{F}_{LC}))),}
    \label{8}
\end{equation}
where $\mathcal{M}_i{(\cdot)}$ means MLP,
$\odot$ refers to the Hadamard product, \emph{i.e.}, the product between elements, and $\delta {(\cdot)}$ represents the activation function, \emph{i.e.}, SiLU~\cite{shazeer2020glu}. 

In this step, the model can utilize the property of Mamba to achieve selectively giving focus to the inputs based on themselves, which can be expressed by rewriting \cref{5}: 
\begin{equation}
\begin{split}
&s_B(\mathbf{F}_{LC}),\ s_C(\mathbf{F}_{LC}),\ s_{\Delta}(\mathbf{F}_{LC}) =\\
&\quad \operatorname{Linear}_B(\mathbf{F}_{LC}), \operatorname{Linear}_C(\mathbf{F}_{LC}),\operatorname{Linear}_{\Delta}(\mathbf{F}_{LC}),
\end{split}
\label{select}
\end{equation}
$s_B(\mathbf{F}_{LC}),\ s_C(\mathbf{F}_{LC}),\ s_{\Delta}(\mathbf{F}_{LC}) $ can decide whether to give the input a larger or a lower weight, thus allowing irrelevant information to be filtered out and relevant information to be retained indefinitely~\cite{gu2023mamba}. 

Though the proposed Mamba Filter can focus on or discard particular inputs to mining useful information,  its architectural constraints in cross-channel interaction mechanisms~\cite{guo2024mambair} require careful analysis, which plays a crucial role in high-precision correspondence learning where cross-channel correlations encode essential geometric constraints. 
Thus, we customarily design a channel-aware module as shown in the right half of \cref{mamba}, denoted as:
\begin{equation}
    \mathbf{F}_{CA}=\mathbf{F}_M\odot W,
    \label{fca}
\end{equation}
where $W$ represents the expressive capabilities of different channels, defined as:
\begin{equation}
    W=\operatorname{Sigmoid}(\mathcal{M}_5(\delta(\mathcal{M}_4(\operatorname{AP}(\mathbf{F}_M))))),
    \label{10}
\end{equation}
where $\operatorname{AP}{(\cdot)}$ denotes average pooling. We further perform Context Normalization~\cite{yi2018learning} on the feature, which is important for its access to context information. Then the \cref{fca} and \cref{10} are optimized as:
\begin{equation}
	\begin{split}
	\mathbf{F}_{CA}=&\mathbf{F}_{CN}\odot W', \quad
\mathbf{F}_{CN}=\delta(\operatorname{BN}(\operatorname{CN}(\mathbf{F}_M))),\\
W'=&\operatorname{Sigmoid}(\mathcal{M}_2(\delta(\mathcal{M}_1(\operatorname{AP}(\mathbf{F}_{CN}))))).
	\end{split}
 \label{11}
\end{equation}

In summary, our initial input feature $\mathbf{F}$ is finalized by CSLB, LGPL, and CAMF to get $\mathbf{F}_{CA}$ with \cref{F_11}, \cref{14}, \cref{8}, and \cref{11}. Next, we use the inlier predictor $\operatorname{IP}{(\cdot)}$~ \cite{yi2018learning} to predict the probability of getting the match to be an inlier, denoted as follows:
\begin{equation}
    \mathbf{p}=\operatorname{IP}(\mathbf{F}_{CA}).
    \label{p}
\end{equation}
Notably, we reorder $\mathbf{F}_{CA}$ with $\mathcal{I}$ to its original order to ensure the output logits are consistent with the input feature. We omit this step in the text for the sake of brevity.
\begin{table}[t]
    \centering
    \resizebox{0.99\linewidth}{!}{
    \begin{tabular}{lcc}
        \toprule
        Method & \multicolumn{1}{c}{YFCC100M~\cite{thomee2016yfcc100m}} & \multicolumn{1}{c}
        {SUN3D~\cite{xiao2013sun3d}} \\
        \hline
        GMS~\cite{bian2017gms}              & 13.29/24.38/37.83 & 4.12/10.53/20.82 \\
        LPM~\cite{ma2019locality}           & 15.99/28.25/41.76 & 4.80/12.28/23.77 \\
        CRC~\cite{fan2021smoothness}        & 16.51/28.01/41.38 & 4.07/10.44/20.87 \\
        VFC~\cite{ma2014robust}             & 17.43/29.98/43.00 & 5.26/13.05/24.84 \\
        PointCN~\cite{yi2018learning}       & 26.73/44.01/60.49 & 6.09/15.43/29.74 \\
        OANet~\cite{zhang2019learning}      & 27.26/45.93/63.17 & 6.78/17.10/32.41 \\
        CLNet~\cite{zhao2021progressive}    & 31.45/51.06/68.40 & 6.67/16.81/31.45 \\
        LMCNet~\cite{liu2021learnable}      & 30.48/49.84/66.94 & 6.84/17.62/33.44 \\
        ConvMatch~\cite{zhang2023convmatch} & 31.69/51.41/68.45 & 7.32/18.45/34.41 \\
        NCMNet~\cite{liu2023progressive}    & 32.30/52.29/69.65 & 7.10/18.56/{\bf 35.58} \\
        MC-Net~\cite{li2024mc}              & 33.02/52.42/69.23 & 7.40/18.49/34.61\\
        DeMatch~\cite{zhang2024dematch}     & 32.98/52.37/69.01 & 7.44/18.66/34.78\\
        \rowcolor{gray!30}CorrMamba (Ours) & {\bf 34.92}/{\bf 54.88}/{\bf 71.59} & {\bf 7.68}/{\bf 18.87}/{34.87} \\
        \bottomrule
    \end{tabular}
    }
    \caption{Results (AUC@5\textdegree/@10\textdegree/@20\textdegree) of relative pose estimation with RANSAC. The best results are marked in bold.}
    \label{tab:relative}
\end{table}
\subsection{Loss Function}
We choose a widely used loss function~\cite{yi2018learning,zhang2023convmatch} as:
\begin{equation}
L=\sum_{l=1}^{\mathcal{L}}\alpha L_{cls}(^{(l)}\mathbf{p},\mathbf{Z})+\beta L_{reg}(^{(l)}\widehat{\mathbf{E}}, \mathbf{E}).
\end{equation}
Herein, $\mathcal{L}$ is the number of layers, 
$\mathbf{Z}=\{z_i\}^N_{i=1}$ encapsulates weakly supervised labels derived via geometric error ~\cite{hartley2003multiple}, $\widehat{\mathbf{E}}$ denotes the estimated essential matrix, and $\mathbf{E}$ stands for the ground-truth. 
$L_{cls}(\cdot)$ denotes a rudimentary binary cross-entropy loss designed for the classification aspect, while $L_{reg}(\cdot)$ is ascertained utilizing the Sampson distance~\cite{hartley2003multiple}:
\begin{equation}
\begin{aligned}
&L_{\text {reg }}(\widehat{\mathbf{E}}, \mathbf{E}) =\\& \sum_{i=1}^N \frac{\left(\mathbf{t}_i^{\prime \top} \widehat{\mathbf{E}} \mathbf{t}_i\right)^2}{\left\|\mathbf{E t}_i\right\|_{[1]}^2+\left\|\mathbf{E} \mathbf{t}_i\right\|_{[2]}^2+\left\|\mathbf{E}^T \mathbf{t}_i^{\prime}\right\|_{[1]}^2+\left\|\mathbf{E}^T \mathbf{t}_i^{\prime}\right\|_{[2]}^2},
\end{aligned}
\end{equation}
where $\mathbf{t}_i$ and $\mathbf{t}_i^{\prime}$ represent two keypoints that constitute the correspondence $\mathbf{c_i}$, and $||v||_{[i]}$ denotes the $i$-th element of $\mathbf{v}$.

\subsection{Implementation Details}
CorrMamba consists of $\mathcal{L}=4$ layers. 
and is optimized by Adam. The learning rate is set to $10^{-3}$ during the first $80k$ iterations, then decaying with a factor of $0.999996$ every step. The batch size is $32$, with $\beta$ starting at $0$ and then $0.5$ after the first $20k$ iterations, while $\alpha$ is fixed at $1$.
We set the threshold for distinguishing between inliers and outliers to 0. See the \emph{supplementary material} for more details.

\section{Experiment}
\label{Experiment}

\subsection{Relative Pose Estimation}
\textbf{Datasets.}
We choose YFCC100M~\cite{thomee2016yfcc100m} to demonstrate our method's capability to learn in outdoor environments and SUN3D~\cite{xiao2013sun3d} to showcase its performance in indoor settings. Just as with the classic methods~\cite{zhang2019learning, zhang2024dematch}, we divide the datasets into training, validation, and testing sets, with detailed explanations provided in the \emph{supplementary materials}.


\textbf{Evaluation Protocols.}
We assess the accuracy of pose estimation by analyzing the area under the cumulative error curve (\emph{i.e.}, AUC) for pose errors across various thresholds (5\textdegree, 10\textdegree, 20\textdegree). Pose error is defined as the maximum of the angular error in rotation and translation. We use SIFT~\cite{lowe2004distinctive} to extract up to $2k$ keypoints and acquire putative matches with the NN method.

\textbf{Baseline.}
In our experiments, we categorize the approaches based on their underlying principles and endeavor to compare a wide array of the SOTA, encompassing traditional approaches (GMS~\cite{bian2017gms}, LPM~\cite{ma2019locality}, CRC~\cite{fan2021smoothness}, VFC~\cite{ma2014robust}) and learning-based techniques (PointCN~\cite{yi2018learning}, PoitnACN~\cite{sun2020acne}, OANet~\cite{zhang2019learning}, CLNet~\cite{zhao2021progressive}, LMCNet~\cite{liu2021learnable}, ConvMatch~\cite{zhang2023convmatch}, NCMNet~\cite{liu2023progressive}, MC-Net~\cite{li2024mc} and DeMatch~\cite{zhang2024dematch}).
\begin{figure}[t]
	\begin{center}
		\includegraphics[width=0.24\linewidth]{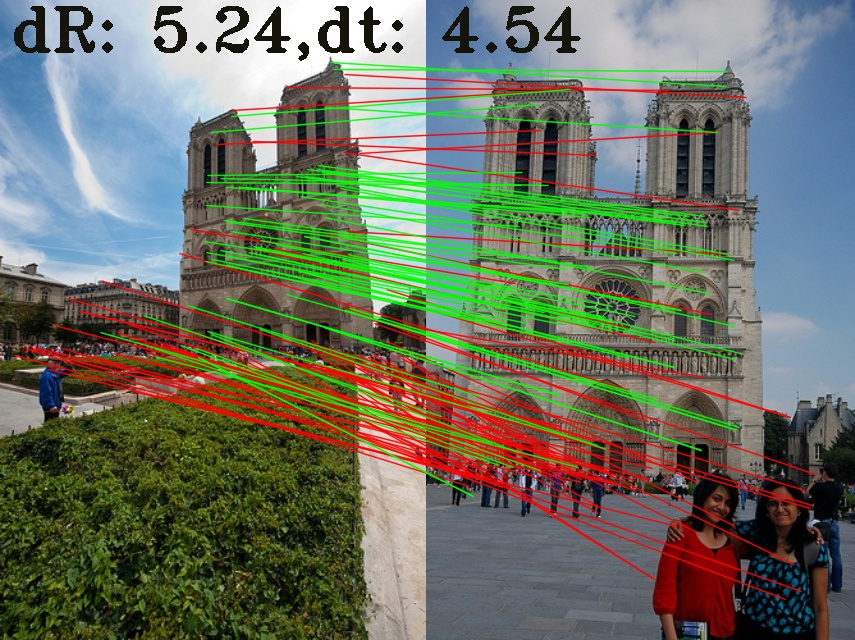}
		\hspace{-0.05in}
		\includegraphics[width=0.24\linewidth]{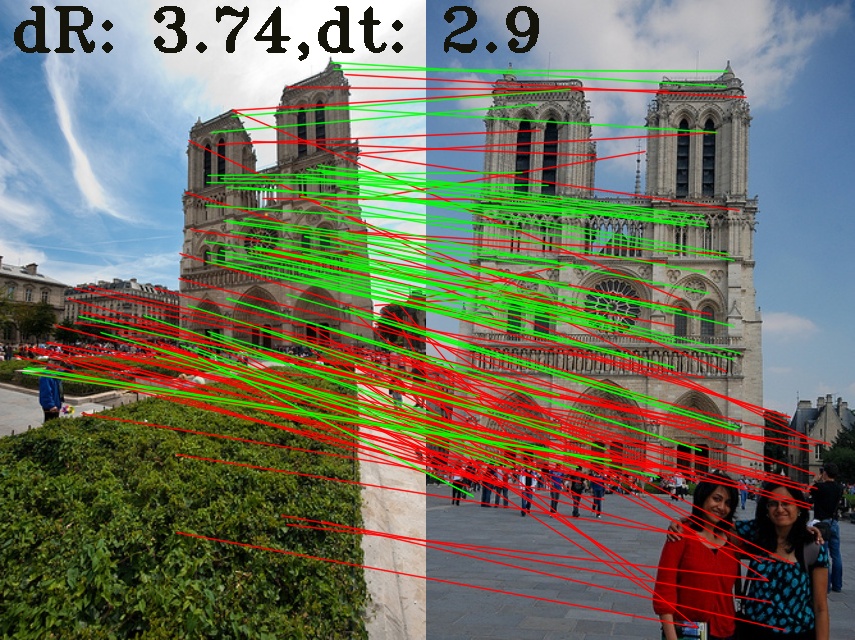}
    \hspace{-0.05in}
		\includegraphics[width=0.24\linewidth]{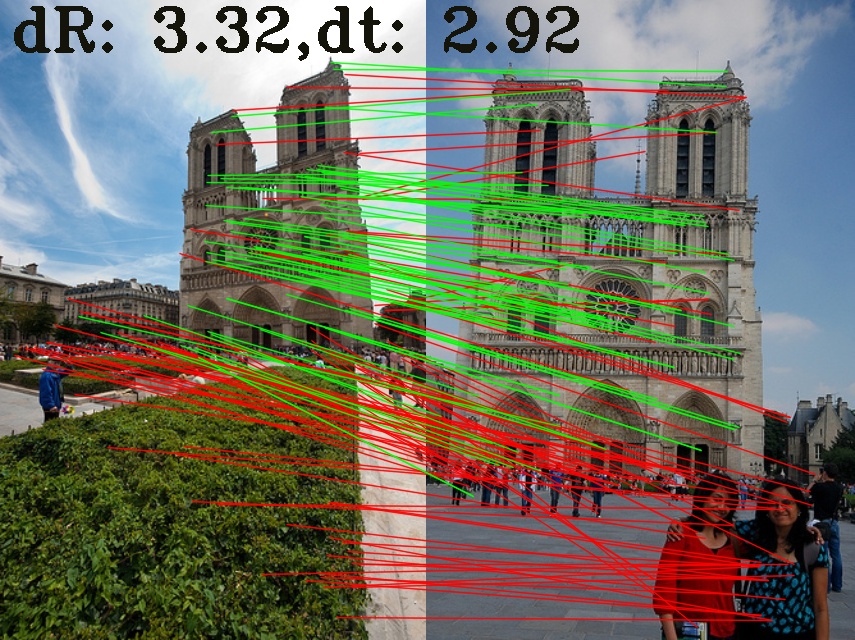}
    \hspace{-0.05in}
		\includegraphics[width=0.24\linewidth]{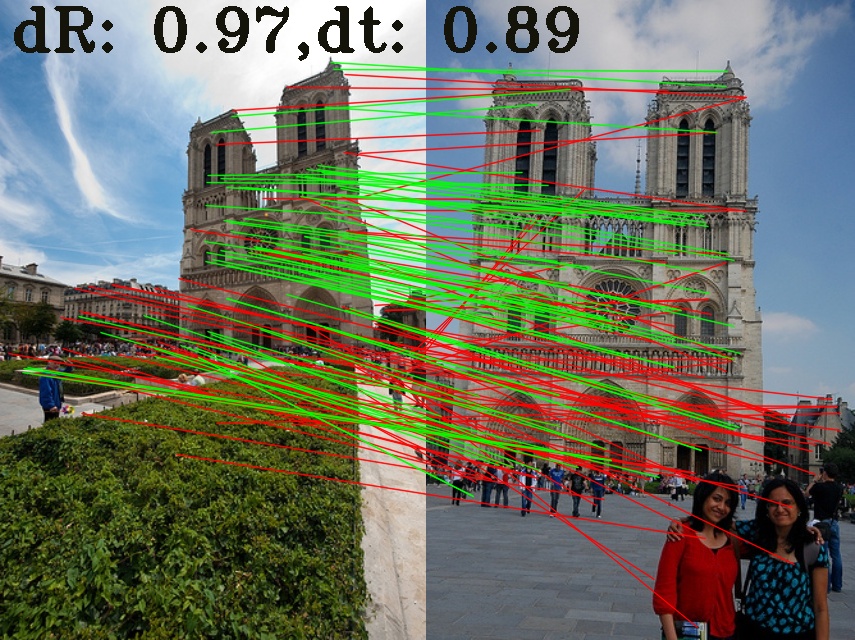}\\
        \includegraphics[width=0.24\linewidth]{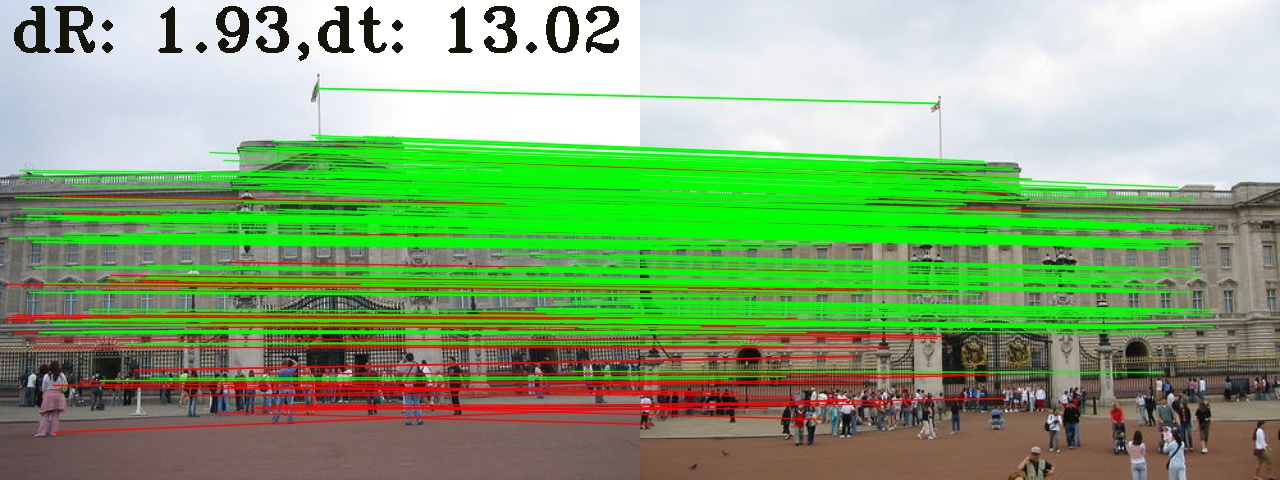}
		\hspace{-0.05in}
		\includegraphics[width=0.24\linewidth]{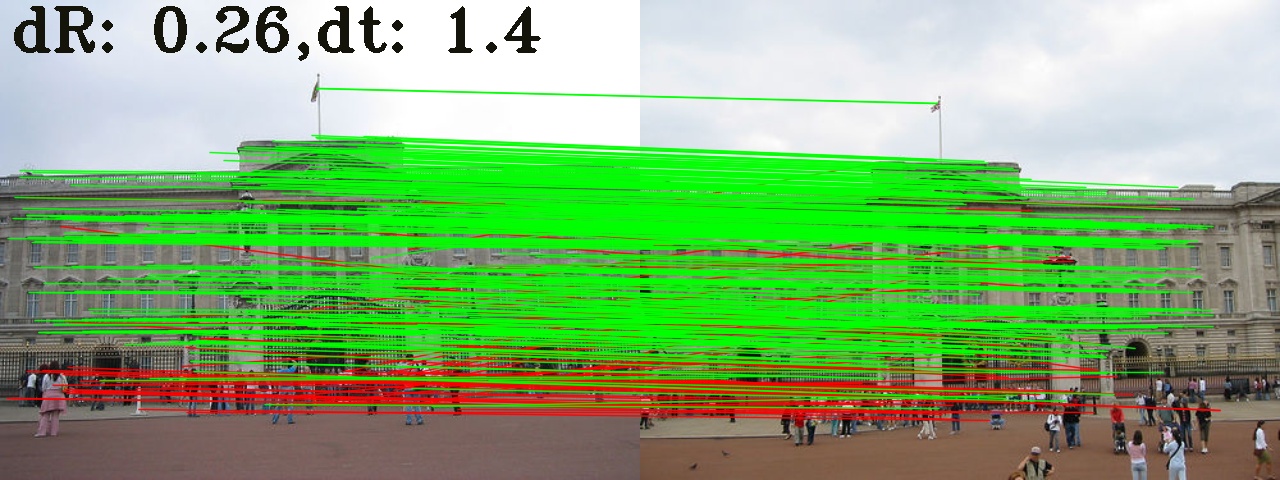}
        \hspace{-0.05in}
		\includegraphics[width=0.24\linewidth]{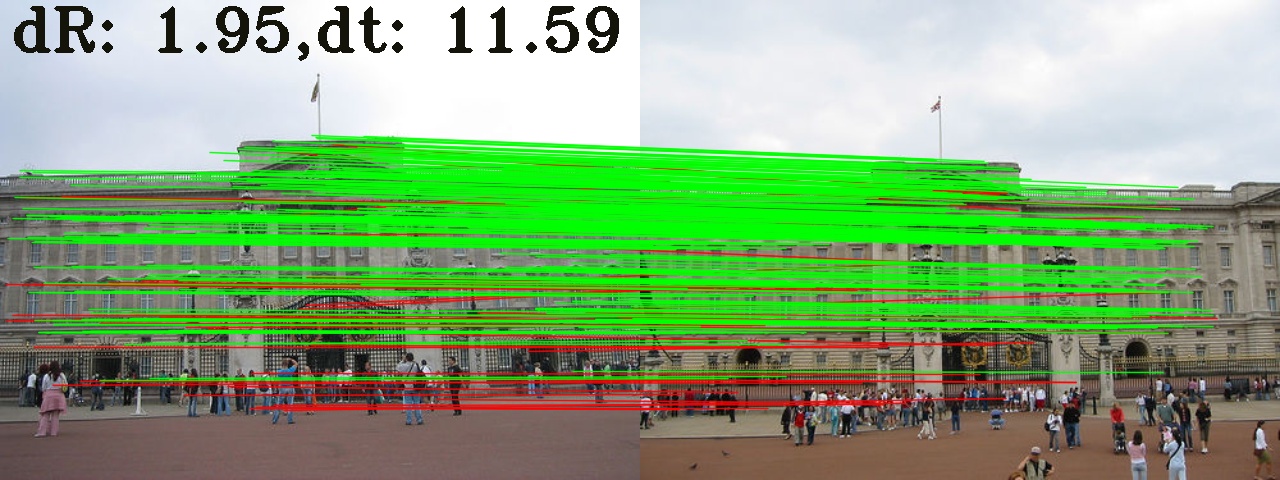}
        \hspace{-0.05in}
		\includegraphics[width=0.24\linewidth]{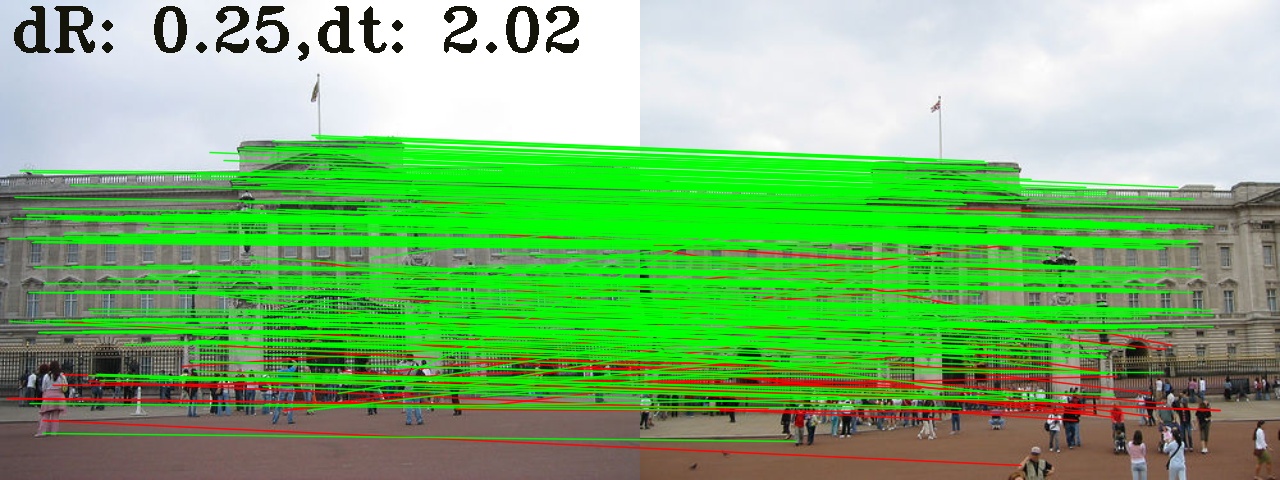}\\
        \includegraphics[width=0.24\linewidth]{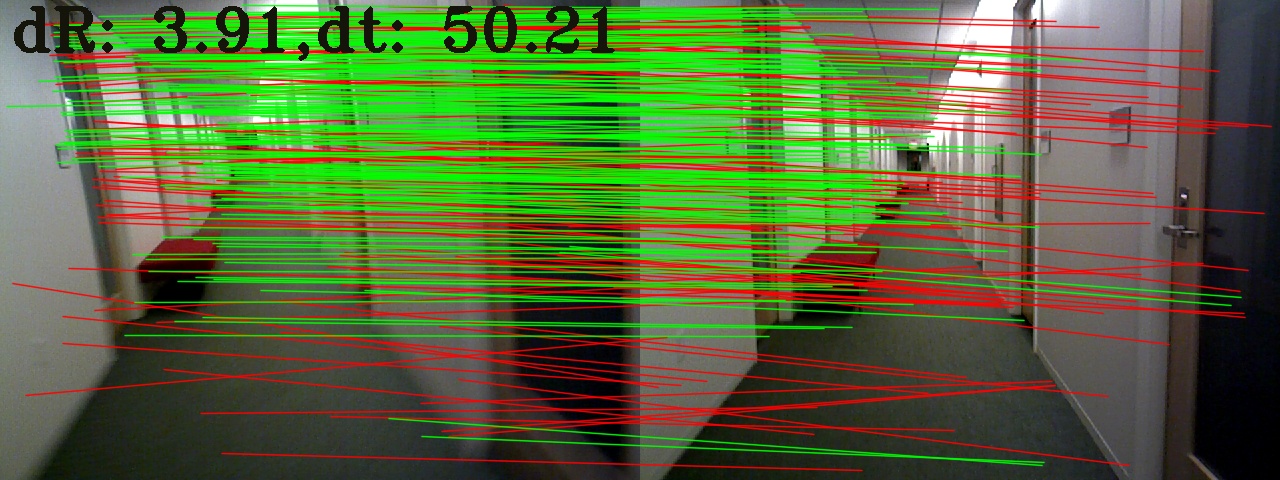}
		\hspace{-0.05in}
		\includegraphics[width=0.24\linewidth]{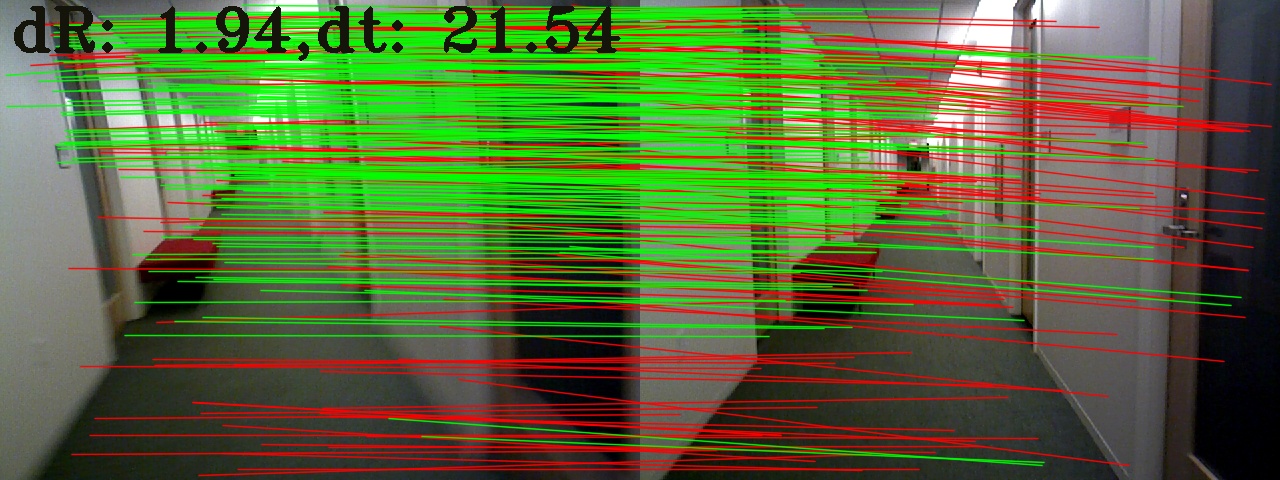}
        \hspace{-0.05in}
		\includegraphics[width=0.24\linewidth]{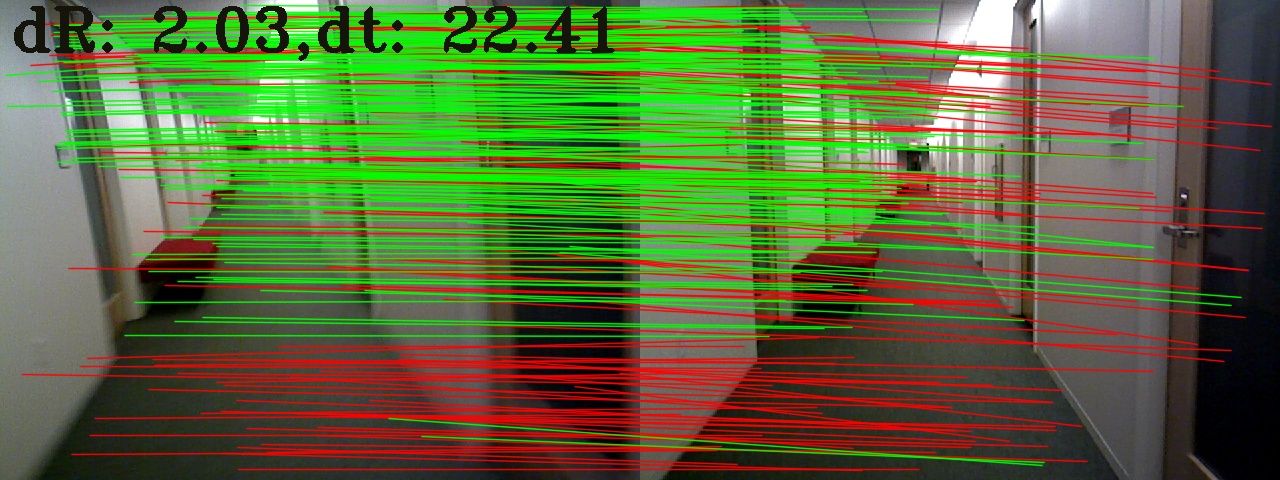}
        \hspace{-0.05in}
		\includegraphics[width=0.24\linewidth]{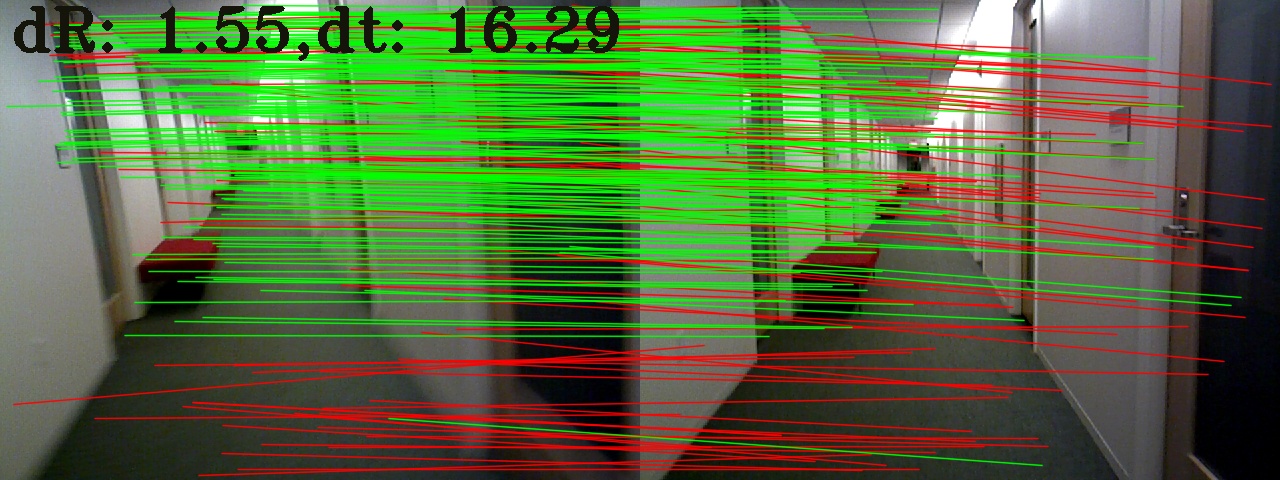}\\
        
        \includegraphics[width=0.24\linewidth]{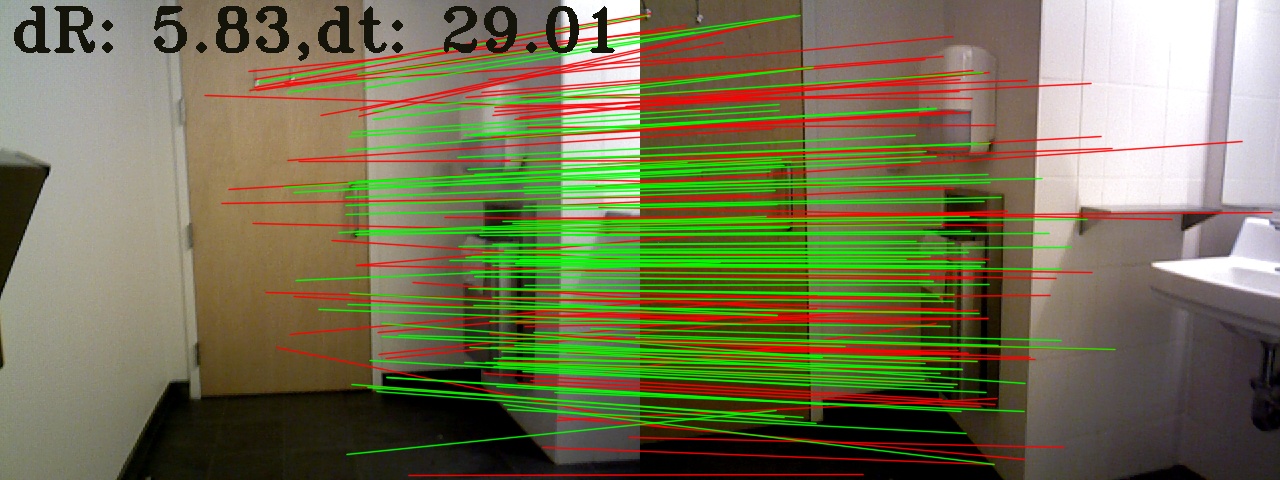}
		\hspace{-0.05in}
		\includegraphics[width=0.24\linewidth]{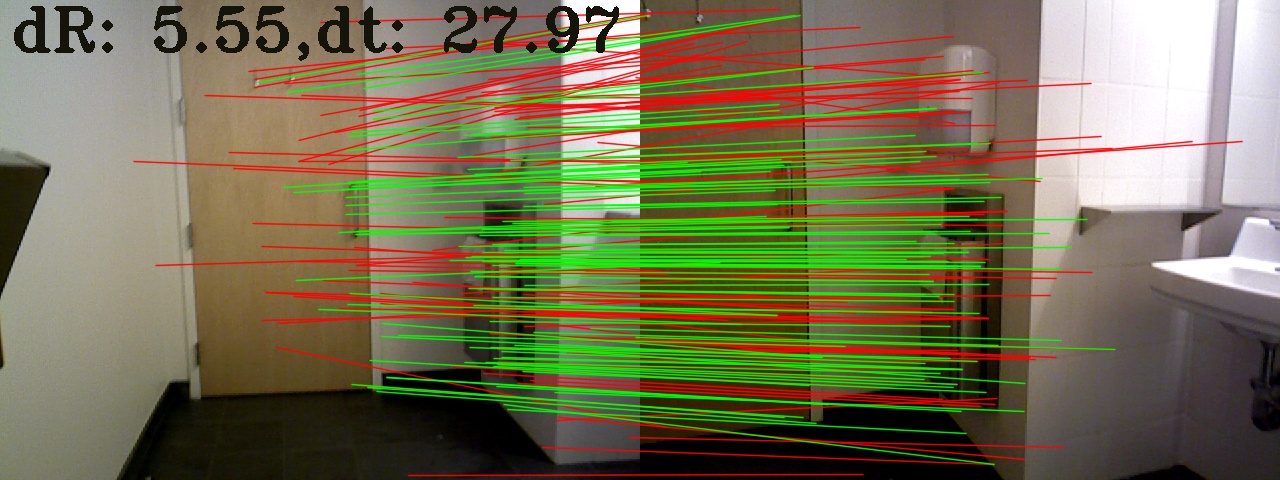}
        \hspace{-0.05in}
		\includegraphics[width=0.24\linewidth]{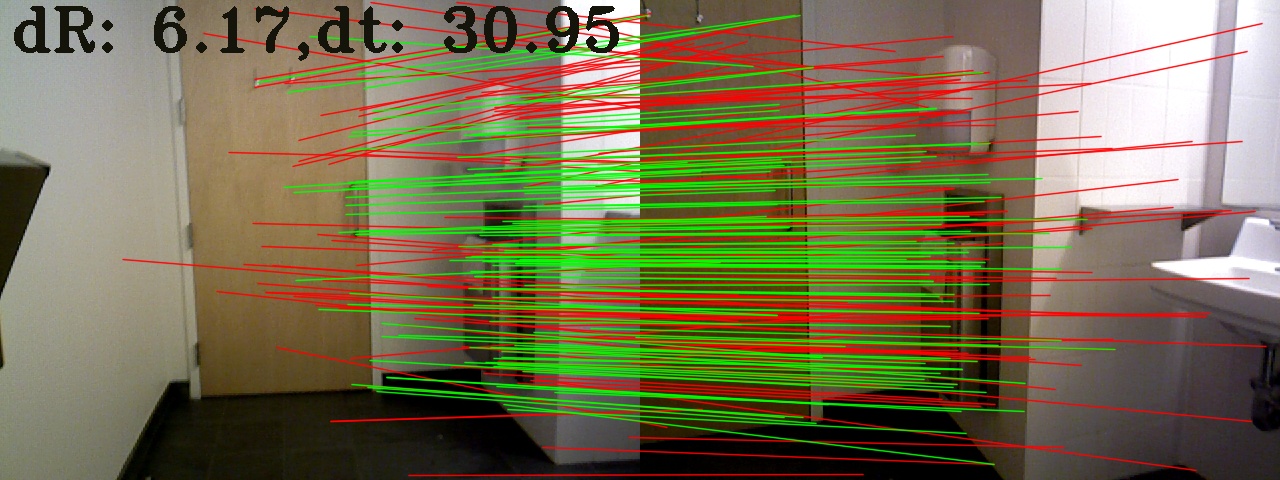}
        \hspace{-0.05in}
		\includegraphics[width=0.24\linewidth]{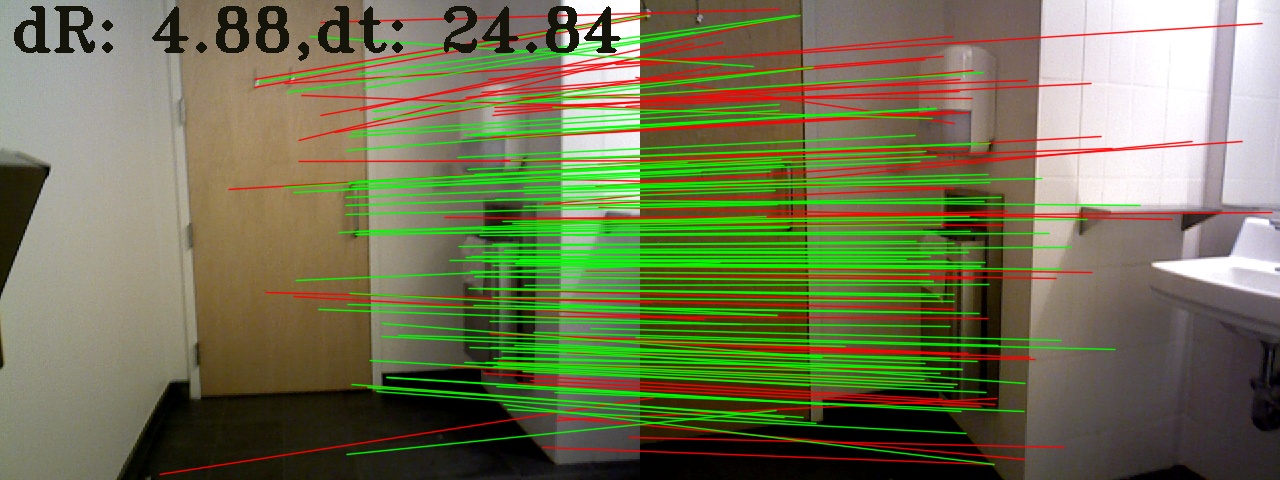}\\
	\end{center}
    \vspace{-0.16in}
    \hspace{0.22in} \small{OANet} \hspace{0.30in} \small{LMCNet} \hspace{0.25in} \small{ConvMatch} \hspace{0.1in} \small{CorrMamba}\\
    \vspace{-0.2in}
	\caption{
		Qualitative illustration of outlier rejection. False matches are marked with red (\textcolor[rgb]{1,0,0}{—}) while correct matches are with green (\textcolor[rgb]{0,1,0}{—}). The relative pose estimation results (error of rotation and translation) are provided in the top left corner of each image pair. Please zoom in for a better view.
	}
	\label{tu_relative}
\end{figure}

\textbf{Results.} \cref{tab:relative} presents the estimation outcomes for both YFCC100M and SUN3D. Drawing upon pertinent literature~\cite{sarlin2020superglue, zhao2021progressive}, we adopt RANSAC~\cite{fischler1981random} as our robust essential matrix estimator. Employing the innovative Mamba architecture for feature matching, we juxtapose our proposed CorrMamba against popular methodologies. 
It is evident that our method surpasses nearly all qualitative assessments compared to the previous SOTA methods. 
Additionally, we provide qualitative insights into outlier rejection, as illustrated in \cref{tu_relative}. Our approach excels at preserving inliers and effectively eliminating outliers, thereby achieving relative pose estimation with minimal rotation and translation errors. 


\subsection{Visual Localization}
\begin{table}[t]
    \small
    \centering
{
\begin{tabular}{lcc}
  \toprule
  \multirow{2}{*}{Method} & Day &Night\\
  \cline{2-3}
  & \multicolumn{2}{c}{(0.25m,2\textdegree)/(0.5m,5\textdegree)/(1.0m,10\textdegree)}\\
  \hline
PointCN~\cite{yi2018learning}       &83.1/92.2/96.2         &69.4/79.6/89.8\\
OANet~\cite{zhang2019learning}    &83.3/92.5/96.6         &71.4/80.6/90.8\\
CLNet~\cite{zhao2021progressive}  &83.3/92.4/97.0	        &71.4/80.6/93.9\\
MS$^2$DG-Net~\cite{dai2022ms2dg}         &84.2/{92.8}/97.0       &\textbf{74.5}/83.7/91.8\\
LMCNet~\cite{liu2021learnable}     &84.1/{92.8}/97.1&71.4/81.6/93.9\\
ConvMatch ~\cite{zhang2023convmatch}   &84.5/92.7/96.8         &{73.5}/83.7/91.8\\
DeMatch~\cite{zhang2024dematch} &{\bf 84.8}/92.7/97.0 & 73.6/{\bf 84.7}/{\bf 94.9} \\
\rowcolor{gray!30} CorrMamba (Ours)    & {\bf 84.8}/{\bf 93.0}/{\bf 97.5} & {\bf 74.5}/82.7/{\bf 94.9}        \\           
\bottomrule
\end{tabular}
}
    \caption{Visual localization results. }
\label{biao_hloc}
\end{table}
To substantiate the practical applicability of CorrMamba, we conduct the visual localization experiment using the official HLoc pipeline~\cite{sarlin2019coarse}.
Engineered to pinpoint the 6-degree-of-freedom (6-DOF) position of query images within a 3D model, this framework serves as the foundation for our assessment. Building upon it, we scrutinize the proficiency of our technique in yielding robust matching outcomes under demanding conditions, including shifts in viewpoint and transitions from daytime to nocturnal illumination.

\textbf{Datasets.}
We leverage the Aachen day-night dataset~\cite{sattler2018benchmarking} to assess the effectiveness of our methodology in visual localization. For a detailed account of the dataset processing, please refer to the \emph{supplementary materials}.


\textbf{Evaluation Protocols.}
As per the authoritative assessments~\cite{sarlin2019coarse}, we present the proportion of accurately localized queries within specified thresholds of distance and orientation. It's noteworthy that we employ SIFT~\cite{lowe2004distinctive} to extract up to $4k$ keypoints from each image. These keypoints are subsequently matched utilizing the NN method to establish putative correspondences. Following this, triangulation is performed on the SfM model using daytime images with known poses. Finally, we leverage correspondence learning for 2D matching and register nighttime query images with the COLMAP framework~\cite{schonberger2016structure}.

\textbf{Results.}
 \cref{biao_hloc} shows the results of visual localization. CorrMamba achieves the best results in most conditions both daytime and nighttime, showing that CorrMamba is able to accurately locate the correct match for scenes with large viewpoint changes and lighting changes. We also report the qualitative results in the \emph{supplementary material}.

\subsection{Analysis}
\label{4_4}

\begin{table}[t]
    \small
    \centering
    {
        \begin{tabular}{lcccc}
            \toprule
            Method & P. (M) & FlOPs. (G) & T. (ms) \\
            \hline
             LMCNet~\cite{liu2021learnable} & 0.925 & - & 227.15 \\
            NCMNet~\cite{liu2023progressive}    & 4.485  & 8.717 & 138.36 \\
            MC-Net~\cite{li2024mc}              & 50.393 & 8.720 & 39.40 \\
            \rowcolor{gray!30} CorrMamba (Ours) & 3.646 & 8.156 & 26.98\\
            \bottomrule
\end{tabular}
}
    \caption{Computational usage.}
\label{computation}
\end{table}
\begin{table}[t]
    \small
    \centering
    \resizebox{1\linewidth}{!}
    {
        \begin{tabular}{cccccc}
            \toprule
            Matcher & Filter & Estimator & @5\textdegree & @10\textdegree & @20\textdegree \\
            \hline
            \multirow{4}{*}{SP + SG ~\cite{detone2018superpoint, sarlin2020superglue}}
            & \ding{55} & \multirow{2}{*}{RANSAC~\cite{lowe2004distinctive}} & 38.06 & 58.38 & 74.67 \\
            & $\checkmark$&                                                  & 39.42 & 60.78 & 76.79\\
            \cline{2-6}
            & \ding{55} &  \multirow{2}{*}{PROSAC~\cite{chum2005matching}}& {39.41} & {59.00} & {74.91} \\
            & $\checkmark$ &  & 40.06 & 60.18 & 75.53 \\
            \hline
            \multirow{4}{*}{SP + LG~\cite{detone2018superpoint, lindenberger2023lightglue}}
            & \ding{55} & \multirow{2}{*}{RANSAC} & 39.42 & 59.69 & 75.89 \\
            & $\checkmark$ &                       & 40.67 & 60.74 & 76.93\\
            \cline{2-6}
            & \ding{55} & \multirow{2}{*}{PROSAC} & 39.93 & {59.82} & 74.21 \\
            & $\checkmark$ &  & 40.76 & 60.43 & 75.76 \\
            \hline
            \multirow{4}{*}{LoFTR~\cite{sun2021loftr}}
            & \ding{55} & \multirow{2}{*}{RANSAC} & 39.80 & 60.03 & 76.07 \\
            & $\checkmark$&                       & 40.72 & 61.12 & 77.04 \\
            \cline{2-6}
            & \ding{55}     &  \multirow{2}{*}{PROSAC} & {42.58} & {58.14} & {70.04} \\
            & $\checkmark$  &                          & 43.04 & 59.47 & 71.89 \\
            \bottomrule
\end{tabular}
}
    \caption{Compatibility with Matchers.}
\label{matcher}
\end{table}

\textbf{Computational Usage.}
The previous experiments demonstrate the effectiveness of our approach on various visual tasks. To further validate the efficiency of our method, we evaluate several performance metrics, including model parameters (P.), floating-point operations (FLOPs.), and average running time per image (T.). These evaluations are conducted using the same test dataset ($4k$ images, each extracted $2k$ keypoints using SIFT~\cite{lowe2004distinctive}). The results, presented in \cref{computation}, indicate that our method offers advantages over vanilla attention-based methods. While LMCNet~\cite{liu2021learnable} significantly reduces model size by leveraging additional library functions, its time consumption is relatively high due to graph construction and matrix decomposition used to solve explicit regularization terms.

\begin{table*}[t]
  \small
  \centering  
    {
    \begin{tabular}{lcccccccc}
    \toprule
    \multirow{2}{*}{Method} & \multicolumn{4}{c}{YFCC100M~\cite{thomee2016yfcc100m}}  & \multicolumn{4}{c}{SUN3D~\cite{xiao2013sun3d}} \\
    \cmidrule(lr){2-5}\cmidrule(lr){6-9}
                                        & SIFT  & RootSIFT  & LIFT  & SuperPoint& SIFT  & RootSIFT & LIFT & SuperPoint    \\ \hline
    PointCN~\cite{yi2018learning}       & 60.49 & 61.55     & 51.84 & 60.96     & 27.49 & 28.35 & 26.10 & 28.53  \\
    PointACN~\cite{sun2020acne} & 64.17 & 65.26 & 57.03 & 63.02 & 27.12 & 27.85 & 26.07 & 28.20 \\
    OANet~\cite{zhang2019learning}      & 63.17 & 65.25     & 57.26 & 63.52     & 26.52 & 27.61 & 26.20 & 27.68 \\
    LMCNet~\cite{liu2021learnable}      & 66.94 & 68.02     & 60.27 & 64.49     & 27.65 & 28.54 & {27.43} & 28.62 \\
    MS$^2$DG-Net~\cite{dai2022ms2dg} & 68.34 & 69.16 & \underline{62.45} & 66.50 & 28.60 & \underline{29.51} & {\bf 27.87} & {\bf 30.55} \\
    ConvMatch~\cite{zhang2023convmatch} & \underline{68.45} & \underline{69.61}     & 62.17 & {\bf 67.36}     & \underline{28.79} & 29.46 & \underline{27.80} & {28.92} \\
    \rowcolor{gray!30} CorrMamba (Ours)& \textbf{71.59} & \textbf{71.81} & \textbf{63.53} & \underline{67.01} & {\bf 29.05} & {\bf 29.56} & {27.42} & \underline{29.32} \\
    \bottomrule
    \end{tabular}}
  \caption{Generalization ability test. We report the AUC@20\textdegree, highlighting the best results in bold and underlining the second. }
\label{generalization_table}
\end{table*}

\textbf{Compatibility with Matchers.}
In the experiments described above, we thoroughly showcase the outstanding capabilities of CorrMamba when utilized with SIFT~\cite{lowe2004distinctive} and the NN matchers. Serving as a versatile backend outlier filter, we assess the performance of CorrMamba on YFCC100M~\cite{thomee2016yfcc100m} for relative pose estimation, with or without various widely adopted matchers. These include SuperPoint~\cite{detone2018superpoint} paired with SuperGlue~\cite{sarlin2020superglue} (referred to as SP + SG), SuperPoint paired with LightGlue~\cite{lindenberger2023lightglue} (referred to as SP + LG), and LoFTR~\cite{sun2021loftr}, while employing RANSAC~\cite{fischler1981random} or PROSAC~\cite{chum2005matching} for pose estimation. For SP + SG and SP + LG, we adhere to the settings of SuperGlue and detect up to $2,048$ keypoints. Evaluation procedures for LoFTR closely follow those outlined in \cite{truong2021learning}. It's worth noting that, in line with recommendations from similar experiments in \cite{liu2021learnable}, we refrain from employing a filtering strategy in each method, opting instead to retain all putative correspondences as inputs. The findings presented in \cref{matcher} demonstrate that as a generalized outlier filtering approach, CorrMamba consistently enhances advanced matchers and can serve as a complementary module in practical applications.

\textbf{Generalization Ability.}
To evaluate the performance of CorrMamba in various scenes and descriptors, we compare several methods trained with SIFT~\cite{lowe2004distinctive} on YFCC100M~\cite{thomee2016yfcc100m}, testing these models on YFCC100M with RootSIFT~\cite{arandjelovic2012three}, LIFT~\cite{yi2016lift}, and SuperPoint~\cite{detone2018superpoint}, as well as on SUN3D~\cite{xiao2013sun3d} 
for pose estimation. For each image, we extract up to $2k$ keypoints using different descriptors, with putative matches generated using the NN method. As shown in \cref{generalization_table}, CorrMamba achieves the best results in almost all cases, clearly demonstrating its superior generalization ability.

\begin{table}[t]
    \small
    \centering
    {
        \begin{tabular}{ccccccc}
        \toprule
        \multirow{2}{*}{N} & \multirow{2}{*} {CSLB} & \multirow{2}{*}{LGPL} & \multicolumn{2}{c}{CAMF} & \multirow{2}{*}{L.F.} & \multirow{2}{*}{@5\textdegree} \\
        \cline{4-5}
         & & & Mamba & C.A. & & \\ \hline
         i. & & & $\checkmark$ & $\checkmark$ & & 30.34\\
        ii. & & $\checkmark$ & $\checkmark$ & & & 29.62\\
        iii. & &  $\checkmark$ & & $\checkmark$ & & 31.37\\
        iv. & & $\checkmark$ & & $\checkmark$ & $\checkmark$ &  31.56\\
        v. & & $\checkmark$ & $\checkmark$ & $\checkmark$ & & 32.42\\
        \rowcolor{gray!30} vi. & $\checkmark$ & $\checkmark$ & $\checkmark$ & $\checkmark$ & & 34.92\\
        \bottomrule
    \end{tabular}
}
    \caption{Results of ablation studies.}
\label{Ablation}
\end{table}

\textbf{Ablation Studies.}
We conduct ablation studies by performing relative pose estimation. The results are presented in \cref{Ablation}. Here, Mamba refers to the vanilla Mamba Filter and C.A. indicates whether a channel-aware design is applied to the Mamba or not. We report the AUC@5\textdegree\ using RANSAC as an estimator on YFCC100M~\cite{thomee2016yfcc100m}. 
By examining i. and v., it is evident that local context plays a certain role in the model's performance. 
Comparing ii. and v., it becomes clear that channel awareness significantly enhances our model. 
Furthermore, iii., iv., and v. demonstrate that Mamba outperforms linear attention (denoted as L.F.)~\cite{wang2020linformer} in feature-matching. 
We ultimately establish the necessity of generating input sequences in a certain order through v. and vi., directly highlighting the effectiveness of CSLB. 

\begin{figure}[t]
	\begin{center}
        \includegraphics[width=0.495\linewidth]{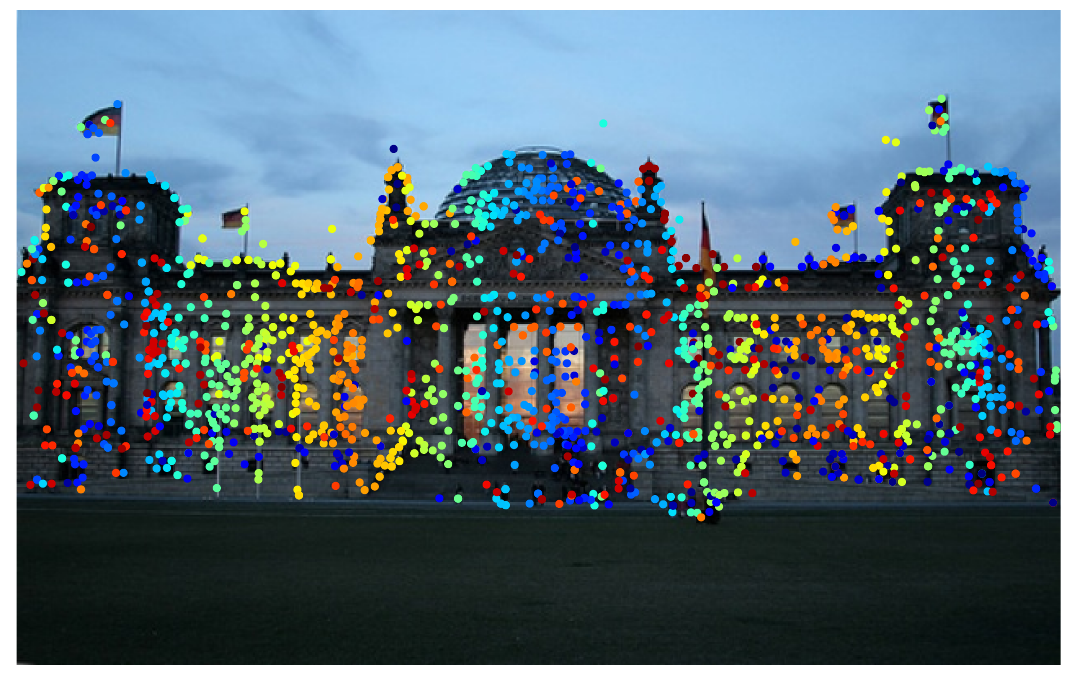}
		\hspace{-0.05in}
		\includegraphics[width=0.47\linewidth]{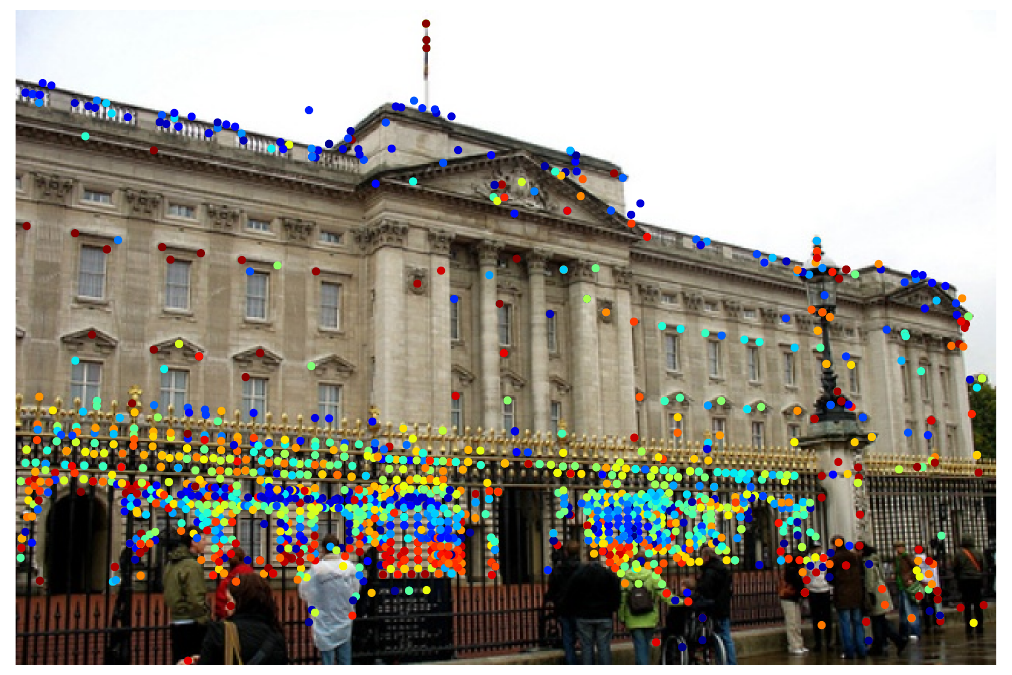}\\
        \includegraphics[width=0.53\linewidth]{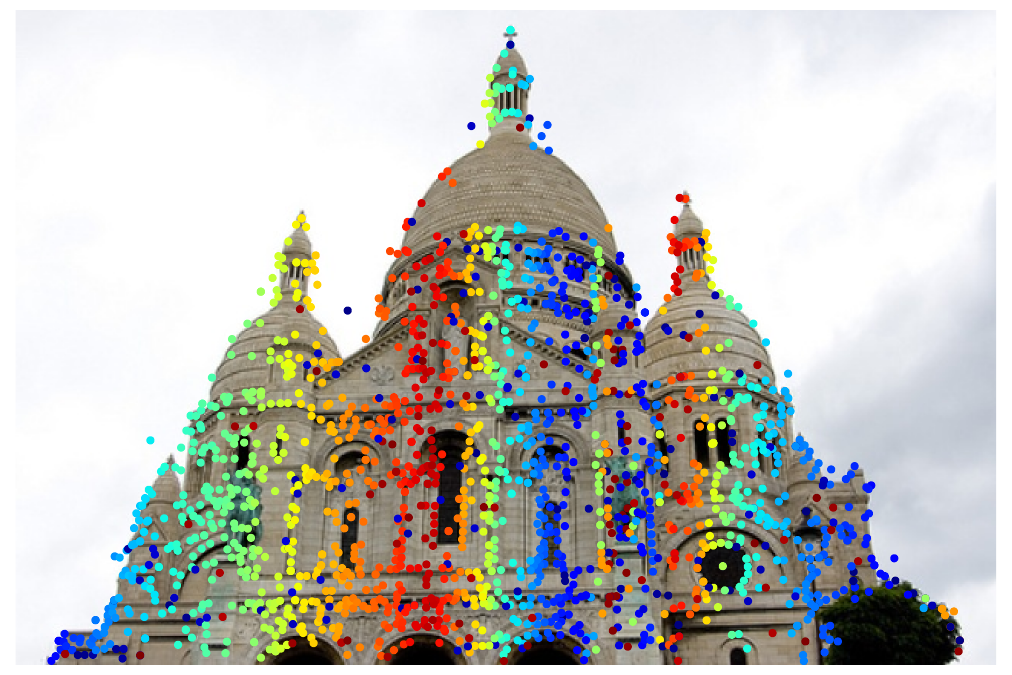}
		\hspace{-0.05in}
		\includegraphics[width=0.435\linewidth]{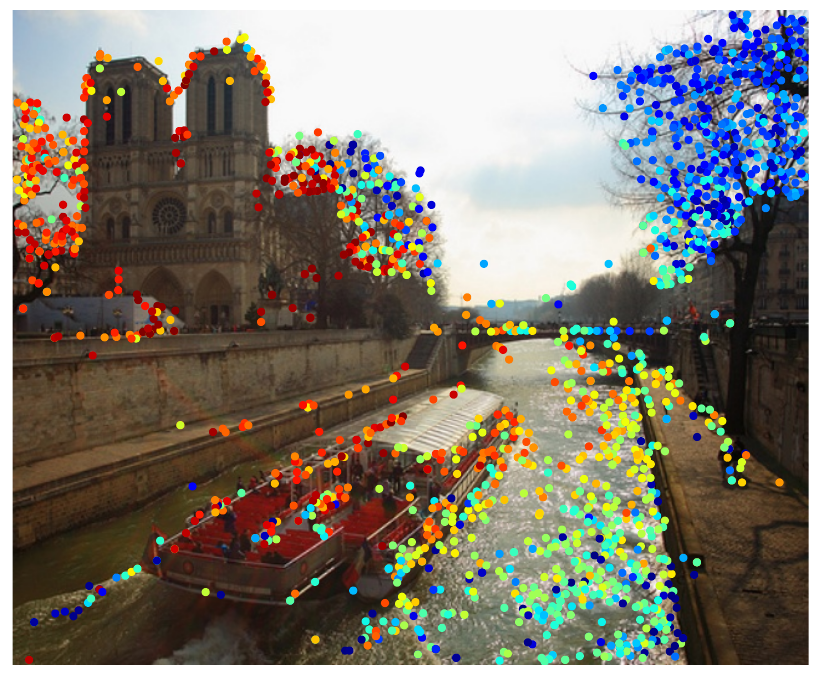}\\
        \includegraphics[width=0.96\linewidth]{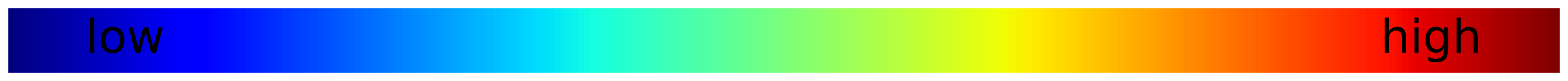}
	\end{center}
    \vspace{-0.2in}
	\caption{Visualization of Sequencialization. We sort the keypoints based on their scores, with red denoting higher rankings, \emph{i.e.} higher scores, and blue indicating lower rankings, \emph{i.e.} lower scores.
}
    \label{tu_sequence}
\end{figure}

\textbf{Understanding Sequencialization.}
 We visualize the learned sequences in \cref{tu_sequence}. Specifically, we gain the predicted scores of different images and rank the keypoints to form a sequence utilizing these scores. It is observed that some of the sequences possess certain characteristics, although these may not be easily definable or observable by humans. 
Firstly, there is regional clustering, where they might form a cluster in a particular area. This could be due to the similarity or consistency of local features. 
Additionally, in some images, features of the detailed parts are ranked higher, while those of the outer contours of buildings are ranked lower. We believe that this is not something that can be easily defined by relying on manual prior knowledge alone, which confirms the advantage of learning sequences in a fully autonomous manner. 


\section{Conclusion}
\label{Conclusion}
In this paper, we design a novel network named CorrMamba for two-view correspondence learning. Inspired by Mamba, CorrMamba can focus on or discard particular inputs. 
Meanwhile, ignoring irrelevant information compresses the context, resulting in redundant information being discarded directly. Moreover, we construct causal sequences in a fully autonomous and differentiable manner to avoid the potential counteraction effects of the clutter and disorder of keypoints.
Comprehensive experiments and analysis prove the superior performance of our proposed CorrMamba.


{\small
\bibliographystyle{ieeenat_fullname}
\bibliography{11_references}
}


\end{document}